# Application Architecture for Spoken Language Resources in Organisational Settings

Rodney J. Clarke, Dali Dong and Philip C. Windridge

Faculty of Computing, Engineering and Technology, Staffordshire University
Beaconside, Stafford ST180DG, United Kingdom
{ R.J.Clarke | D.Dong | P.C.Windridge }@staffs.ac.uk

**Abstract**

Special technologies need to be used to take advantage of, and overcome, the challenges associated with acquiring, transforming, storing, processing, and distributing spoken language resources in organisations. This paper introduces an application architecture consisting of tools and supporting utilities for indexing and transcription, and describes how these tools, together with downstream processing and distribution systems, can be integrated into a workflow. Two sample applications for this architecture are outlined- the analysis of decision-making processes in organisations and the deployment of systems development methods by designers in the field.

## 1. Introduction

Communication is an increasingly important aspect in the study and practice of organisations and information systems. However, in many respects spoken language resources in organisations are entirely overlooked- devalued as 'just talk'- in favour of written artefacts. Part of this is due to the fact that speech appears to be more immediate than the 'value added' deliberations associated with writing. Writing is not simply speech written down, rather spoken and written language are literally different *kinds* of meaning-making systems. Both types of language resource are complex although in different ways; written language is lexically complex while spoken language is grammatically complex (Halliday 1985; Hasan in Halliday and Hasan 1985). Given that the language system is more richly developed and more fully revealed in speech (Halliday 1985), it is not surprising that spoken language resources are often central to organisational activities. In fact it is the grammatical complexity of speech, which makes it significant for organisations, and provides interesting possibilities for the analysis of organisational processes. These characteristics of spoken language resources also necessitate special techniques and technologies, to acquire, transform, store, process, and distribute spoken language artifacts and metadata. An open-source application architecture is described in this paper, which has been developed to explicitly deal with the complexity of spoken language resources in organisational settings. Uses of this architecture are illustrated in the domains of knowledge/skills transfer and decision analysis.

## 2. Challenges of Spoken Language in Organisations

The application architecture described in this paper was developed in response to the unique challenges of spoken language in organisational settings. These challenges include the necessity to overcome bottlenecks associated with the transcription process; giving due consideration to issues of access, control, power and representation; increasing the reporting frequency from researchers to clients; managing corpora; supporting a range of processing options for spoken language; and controlling the dissemination, security and privacy of artifacts and metadata. These challenges are generally ignored, non-existent or largely avoidable in more traditional areas that use spoken language resources. In this section, we concentrate on the first three of these challenges as they have proved to be of immediate concern in our studies.

### 2.1 *Throughput Issues*

Transcription is the process of representing speech as writing, and coding is the process of adding relevant information to the transcript. Transcription and coding processes are relatively difficult, time consuming, and therefore costly. Our experience suggests that the production of meeting transcripts can take four to five times the length of the observational record. The time taken to transcribe and code a stretch of spoken language depends on a number of factors. The purpose for the transcript determines the precision with which utterances will need to be transcribed, and the various types of codes that may be applied to them. In some cases, a simplified transcript may suffice, for example a so-called *flowing original* transcript is sufficient for subtitling a web cast. In other cases, like the detailed studies of decision-making during systems development, the transcription will likely need to be exacting and a broad selection of different codes may be employed. Familiarity with transcription standards and tools will have an effect on transcription throughput. Also throughput is improved if the transcriber was present at the social occasion when the spoken language recording was made. Transcription speed and accuracy are improved if the transcriber is transcribing in their first language.

### 2.2 *Access, Control, Power and Representation*

Spoken language in organisations is inextricably linked to issues of access, control, power and representation. The methodology informing the design of this architecture and the approach we have used during our case studies is *empowerment participation* (Cameron et al 1992). Here the participants determine what gets recorded and subsequently transcribed. The result is that a transcript may well have intentionally omitted sections and the resulting artifact is referred to here as a *partial transcript*. The apparent disadvantage for analysts studying organisational processes is the loss of speech excluded from the observational record by clients. These omissions might be justified in terms of material being considered as 'sensitive or political', as 'commercial in confidence', or as the 'intellectual property' of consultants and other third parties. Interestingly, not having full observational records during spoken language occasions is not necessarily a major setback. For example, decision-making is a social process and analysts never have access to all decision-making occasions- these are unpredictable even for the participants themselves. Trust can develop between clients and analysts during case studies placed on the firm basis of empowerment participation.

2.3    *Increased Reporting Frequency*

Given the throughput issues and the raised tensions associated with acquiring spoken language in organisations previously discussed, there can be considerable pressure placed on research teams to demonstrate that these resources once acquired are being put to good use. Moreover, in order to promote trust with the client organisation, the analysis team should provide regular feedback while ensuring that this does not itself become onerous. These apparently contradictory requirements forced a reassessment of how best to deliver spoken language artefacts, which informed the development of some key components of the architecture. Spoken language records can be tagged in a process similar to that employed by professional video logging systems like Virage (2003). Qualitative tagging or *indexing* a given stretch of spoken language takes only a fifth or a quarter of the time required to transcribe it. Nor is this an additional time cost as the judicious selection of indexes can directly contribute to the coding of a transcript. Visual reports can be automatically created from the resulting indexes showing, for example, the topics covered during a meeting. These reports may be issued as deliverables to clients. Rapid reporting of this kind has the added advantage of improving the perceived professionalism of the researchers, promoting trust and confidence between parties- a necessary first step to promoting client ownership and involvement with the research (Clarke 2003).

## 3.    Description of the Architecture

The application architecture consists of four main components together with a potentially large range of processing and distribution options (one of each type are described here) together with several supporting utilities, described in sections 3.1 and 3.2 respectively. These components are designed to be integrated with other third party routines that provide specialist services. The file structure used by the architecture is described in section 3.3. The workflows between the components and support utilities of the architecture, and relevant third party software are described in section 3.4.

3.1    *Components*

The basic components of the architecture include the Digital Media Loop, Indexer, and TransCoder applications together with a lightweight XML file system. These basic components can be augmented with various down stream processing and distribution options. Two additional components that have been used in our own case studies are an experimental natural language processing environment called SemLab, which utilises systemic functional linguistic theory, and a web portal assembled from Post Nuke modules for implementing membership, security, content management, publishing and distribution services.

*Digital Media Loop*: Successful indexing, transcription and coding require tools to handle observational records, possessing features unavailable in typical media player applications. The Digital Media Loop is an application that is designed to support transcription and coding and is built on top of media player functionality provided by the BASS Audio Library (2003). Small sections of the observational record need to be repeatedly played back or *looped* in order for spoken language features to be fully revealed. The starting of the loop, its *duration* or length, and the amount by which the loop is automatically advanced or *offset* to progress to the next section of the observational record, can be changed at any time. The Digital Media Loop can be used to isolate, mark, display, replay and scale a visual representation of the audio waveform. These features assist in resolving acoustic features. The first time an audio file is loaded into the application the waveform display characteristics are calculated and stored. This pre-processing enables the application to very rapidly display the waveform on subsequent accesses of the observational record. Timing information from the media is used to register indexing, transcription and coding information.

*Indexer*: One of the major reasons for creating the application architecture was to support partial transcription- transcripts can be assembled in pieces. Modelled on the use of tagging technologies for event broadcasting, where large amounts of media resources are created and need to be managed, the Indexer enables spoken language records to be indexed prior to transcription and coding. The indexes are in the form of a graphical notation called a *system network* (see Eggins 1994). Analysts indexing spoken language records may create, delete, amend, store and retrieve these system networks. The options within the networks and their topologies can be altered at any time and these alterations are version controlled. Time stamped selections from systems networks can be stored as comments within the transcript, and add data to the Spoken Language Architecture (SLA) XML document shared between the Indexer and the transcription tool (see below). Graphical Reports can be produced from the Indexer showing the relative completeness of the indexing of the observational record and the relative and absolute locations of codes selected by analysts.

*TransCoder*: The complexity of spoken language resources meant that it was important to adopt appropriate transcription and coding standards. Based on previous research (Clarke 1992, 1995), the authors selected Codes for the Human Analysis of Transcripts (CHAT). It is a well-defined standard, robust and extensible coding scheme, developed with computer processing in mind. CHAT was developed by Brian MacWhinney and Jane Walter at the CHILDES (*CHI*ld *L*anguage *D*ata *E*xchange *S*ystem) Research Centre, at the Department of Psychology, Carnegie Mellon University (MacWhinney 1995, 2003). Perhaps because of its initial application in early language acquisition and usage, CHAT is extremely useful at both the phonetic and phonemic transcriptions of recorded speech. All CHAT transcripts have a common structure, as shown in Figure 1. The body of the transcript consists of utterances taken by speakers called *mainlines* which are signaled by an asterix followed by a three character participant identifier. Beneath each mainline are zero or more *dependent tiers*, which are used to represent ancillary information. These are signaled by a percent symbol followed by a three-character code that specifies the kind of information contained in the tier. Many types of dependent tiers exist and the use of a given type will be based on the kind of study being undertaken.

TransCoder is designed to transcribe and code spoken language texts using the CHAT standard and has a number of differences over existing CHAT editors (see for example CED described in MacWhinney 1995). While CHAT is relatively easy to learn, its formal syntax means that coding errors are easily made. In recognition of this, a library of computerised routines called CLAN, has been developed to analyse and validate CHAT transcripts (MacWhinney 1988, 1). Arguably, the most important of these routines check the syntax of the transcripts. The design of TransCoder is based on different rationale. The interface supports the process of building syntactically correct CHAT at the point where transcription and coding takes place, rather than after the transcript is complete. Using TransCoder, the analyst is only ever exposed to examples of well constructed and syntactically error free transcript files. This reduces the learning curve for writing valid CHAT transcripts- an approach advocated in the more general case of programming (see Kernighan and Plauger 1981). The graphical user interface, promotes more intuitive operation by not requiring users to remember large numbers of rules and codes. The display of transcripts can be limited to specific types of information. Planned functionality will include the ability to display and hide or collapse and expand entire sections of transcripts- including all utterances made by one or more participants or all utterances relating to particular topics. Unneeded or unwanted transcription and coding options can be removed which promotes ease of use for transcribers with little experience. As the Indexer and TransCoder share the same XML Schema (Clarke et al 2003), information from the Indexer is directly associated with the relevant dependent tiers.

*Storage*: The choice of how to implement storage services for collections of transcripts- referred to as corpora- depends on its size and to a lesser extent on the downstream processing technologies that will be used (for example presentation services, distribution services, natural language processing) and also how extensively one is prepared to use XML technologies. Several different storage technologies were investigated, including hybrid XML-related technologies (Bonneau et al 2003, 234-238), the Xindice native XML database from Apache (Bradford 2003), and MySQL (DuBois 2000). Because our corpora are relatively small it was decided to create a simple XML file system within .NET. The .NET environment provides access to a number of classes that implement XML features like DOM, sequential access, XPath for queries and retrievals, and XUpdate for batch modifications. Using .NET means that bundling as a standalone package would require no configuration on the target machine.

*Distribution & Processing:* Spoken language resources can be acquired, transformed and stored with the applications described above. Other technologies are required to further process these resources. To distribute Spoken Language Resources and other communicative artefacts, our group has been involved with assembling a PostNuke (2003) based portal to support a virtual research community. The following functionality is supported (1) *About this site* including Help/FAQ, Stats, Top Lists, Recommend Us, and Feedback/Contact; (2) *Communication* including News and Archive, Submit News, Reviews, Wiki, Forums, web Links, Up/Downloads; (3) *Documentation* including Conference Papers and Journal Articles; Courses, Lecture Notes and Tutorials; Talks, Seminars and Presentations, Books, Knowledge Base, Gallery; (4) *Members* options including Members Lists, Lists of Contacts, Digital CVs and a dynamic (5) *User's Menu* including Home, Advanced Search, Sections and Topics functions.

Our group has also been involved in developing a set of XML-based NLP tools called SemLab that utilise concepts from Systemic Functional Linguistics (Halliday 1885; Eggins 1994). One of the applications currently under investigation includes the extraction of indexical lexical items from transcripts that specifically indicate the kinds of social actions and activities that are being described in the text. Indexical lexical items can be arranged into a so-called field taxonomy that can function as a kind of ontology for groups of users.

### 3.2 *Support Utilities*

The Spoken Language Architecture is designed to be open, to the extent that third party systems could be used in conjunction with the components described here. For example, it makes sense to use the Digital Media Loop in conjunction with various kinds of digital audio processing (see B in Figure 3), or to use speech recognition to support the operation of the Indexer or TransCoder (see F and K respectively in Figure 3). Apart from these optional elements, there are three supporting utilities that are continually used when dealing with spoken language resources in organisational settings. A *Contact Management* utility enables Contacts details about participants in transcripts to be stored and linked into transcripts (see Figure 3). It contributes information used to fill out Constant Headers in CHAT transcripts like who was present (*@Participants*), their birthdays and ages (*@Birth*, *@Age*), their socio-economic status (*@SES*), and genders (*@Sex*). When information on a contact is changed it causes a new entry for that contact to be created, leaving the old details intact. In this way a contact history is built up; preserving the integrity of the information from previous social occasions in which the contact participated. A *Material Setting Description* utility enables the entry of information about the physical sites, places, and spaces where spoken language occurs. It contributes information used to fill out Changeable or Repeating Headers in CHAT transcripts like the specific circumstances, activities, and physical environment (respectively *@Situation*, *@Activities*, and *@Room Layout*). The *Resource Logger* utility is used to record instances of, and information about, language artefacts and other media collected during the process of analysis in an organisation. These utilities provide specific kinds of information during indexing, transcription and coding- although they may be replaced by more comprehensive commercial systems if the requisite outputs are available in a form that enables them to be imported into the architecture.

### 3.3 *File Structure*

In organisations, social occasions may generate, or be associated with, various multimedia files including text, image, audio and video. In longitudinal studies a number of social occasions may be analysed as part of a single project- for instance a series of meetings organised around a particular issue. Social occasions may also be considered as part of more than one project. Each spoken language occasion can result in at least one audio and/or video file as well as a number of miscellaneous files, for example, an agenda, any circulated documents, notes, image files, and minutes. Analysis of these artifacts results in indexing information and/or a transcript or transcript section- this last possibility would be marked by an *@New Episode* header in a CHAT transcript (see MacWhinney 1995, 18).

The file structure for the Spoken Language Architecture is shown in Figure 2. A *Project Index* is located at the root of the data structure and it lists all the available projects and their locations. Each project has its own *Project File*, containing information about the social occasions specific to that project. It contains a list of all related social occasions stored as links to one or more CHAT files. In this sense it is similar to the *Documentation file* that provides information about a CHAT corpus (see MacWhinney 2003, 19-20). Project Files also contain an inventory of associated text, image, audio, video and other media resources relevant to these social occasions. The CHAT File holds information as well as associated resources stored as a collection of location addresses, relevant to a social occasion. As a social occasion and its associated media may belong to more than one project, a many-to-many relationship can exist between a CHAT file and a Project File. Other data may be provided from relevant data stores, for the sake of brevity only Contacts, Places and Resources data from the Contact Management, Material Setting Description, and Resource Logger support utilities are shown in Figure 2.

Traditionally CHAT transcripts are stored as plain ASCII files, although the standard is also available in the form of an early XML schema. However for this architecture, we developed and used our own CHAT XML schema in order to overcome a number of disadvantages with the existing schema (see Clarke et al 2003). As our version of the CHAT XML schema is shared between TransCoder and Indexer, the specific content of some of the general coding dependent tiers may actually be sourced from either application.

3.4    *Workflow*

The Transcription process starts with the use of the Digital Media Loop to assist in repeatedly playing an audio stream (A). The operation of this component may be supplemented by using third party audio processing including for example routines for pitch contours, spectrograms, and noise reduction (B). The Digital Media Loop can be used in conjunction with the Indexer. The Indexer can request the Digital Media Loop to provide an excerpt of the observational record (C) while providing it and other major components with timing information (D). The Contact Management utility provides information on participants that is shared with TransCoder (E). Experiments are currently being conducted into the use of off-the-shelf speech recognition to more quickly tag segments of the observational record and to add new options to the tag sets themselves (F). The Indexer can be bypassed entirely allowing the TransCoder to request (G) audio data and receive data and timing information (H). A typical configuration has TransCoder (I) and Indexer (J) exchanging information. Given that both components share the same XML schema, they both contribute data to the developing transcript. Speech recognition technologies can be gainfully employed for entering utterances on mainlines prior to coding (K). The Material Setting Description utility allows analysts to enter information concerning places and locations (L). Both the Indexer and TransCoder use a lightweight XML file system for data retrieval (M) and storage (N). Other types of information are stored in the file system including that provided by the Resource Logger utility which is used to store records about documents and other media gathered during the analysis process (O). The architecture (P) can be used in conjunction with other systems to provide sophisticated natural language processing operations for example SemLab (Q), or to organise secure publishing to interested virtual communities using for example a Portal (R).

**Figure 1**: Structure of a simple CHAT Transcript

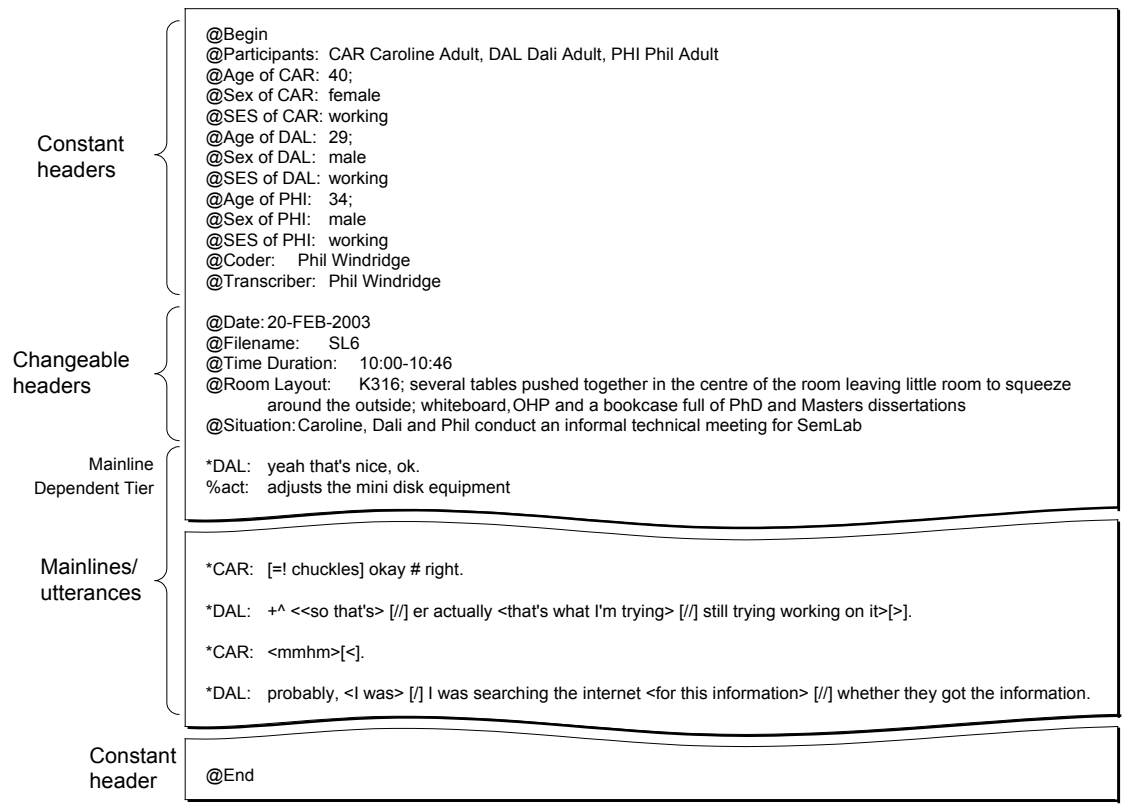

**Figure 2**: File Structure shared by the main components of the Application Architecture.

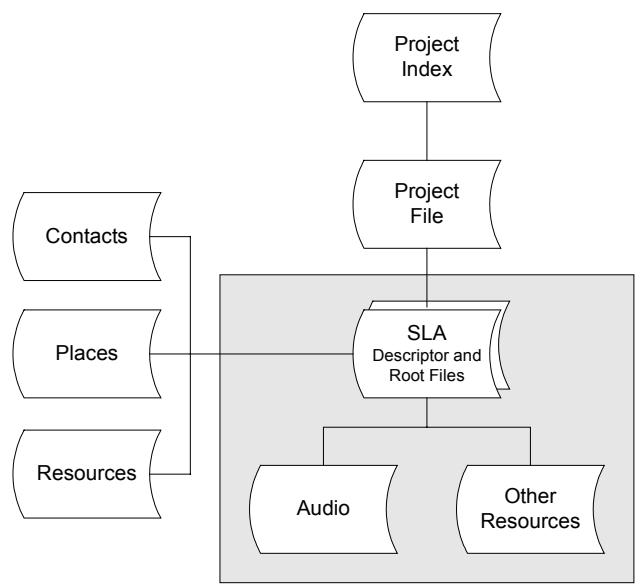

**Figure 3**: Application Architecture for Spoken Language Resources emphasising components and workflow.

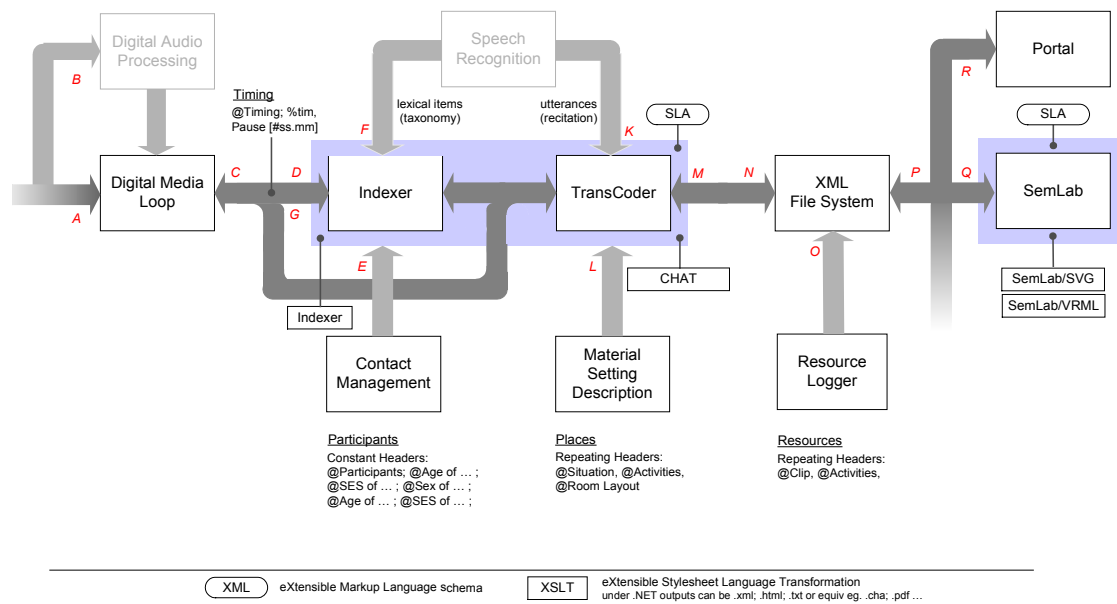

## 4. Sample Applications

The application architecture described above has been developed in response to needs emerging from two very different research projects, the first project involves testing experimental system development methods in workplaces, see section 4.1, and the second project involves understanding decision making in organisational contexts, described in section 4.2.

### 4.1 *In-field Deployment of Systems Development Methods*

Systems development methodologies utilise specific kinds of social occasions, including for example analysis interviews, structured walkthroughs, usability sessions, to explore, propose, and debate development activities and issues. These social occasions provide the opportunity to gather qualitative data including diagrams on whiteboards, video as well as spoken language. A literature has evolved around the rationale for, and acquisition of such multimodal resources within requirements analysis and engineering, workplace studies and ethnography, collaboration technologies in the form of conferencing and voting systems, and the study of interaction at work- see for example Luff et al (1994) and Hughes et al (2000). When it comes to understanding how newer or unproven methods perform in field, there are relatively few approaches available, however a detailed analysis of spoken language provides the means of exploring the performance of the methods and the designers who deploy them. The application architecture for spoken language resources would be usefully applied in these kinds of studies. A detailed linguistic analysis of the speech employed during these occasions would enable the power effects associated with the methods and the development process to be clearly identified. Standard linguistic methods are available for determining- directly from spoken language

resources- the social actions and activities taking place, the role relationships between participants engaged in the application of methods, and the role that language is playing in these interactions. By examining the rhetorical structures used by method advocates, it is relatively easy to distinguish between designers who are actually responding to the client's problems or simply articulating solutions to previously solved model cases. Potentially, the way in which a development method is deployed- the conduct of practice- can be improved by studying how the interaction between method advocates and clients evolves.

4.2     *Analysis of Decision-Making Processes*

Decision-making activities are central to organisations, but the decisions, actions, and issues that comprise them are complex and often require 'insider' knowledge. Decisions, actions and issues may appear to be unrelated to or incompatible with each other, irrelevant to various stakeholder groups, obscured by other related activities, or missing entirely. No single identifiable language units exist to mark the existence of a decision, action or issue- therefore qualitative analysis is difficult and unreliable. The application architecture developed here is being used to collect spoken language in decision-making contexts, which then enables a model of language to be applied to identify various resources used to mark the existence of a decision, action or issue. Decision-making appears to be communicative, contextual, and socio-semantic. The language of decision-making differs from nation to nation and organisation to organisation. Within a given organisation, different social groups exist which share similar institutional, social, economic, and historical experiences- their members can be distinguished from others through their use of situational languages (technically known as Registers, see Halliday 1985). Our investigations have suggested that members of social groups draw on a range of grammatical resources to signal the existence of Issues, Actions and Decision. Issues seem to be signaled using Thematic and Information structures in transcripts, actions appear to be signaled by the use of Material Processes (for example 'real' or action), while decisions may be signaled by Verbal and Mental Processes (for example verbal action involving 'saying', 'thinking' or 'feeling')- if they are evident at all.

5.      **Conclusions and Further Research**

Spoken language resources play a central although often unacknowledged role in organisations. Spoken language is structured differently to written language and exhibits grammatical rather than lexical complexity. Additionally spoken language presents unique challenges for organisations including but not limited to potential throughput bottlenecks during transcription and coding, issues of access, control, power, and representation, as well as the necessity for increased reporting frequency concerning the status of spoken language deliverables. This paper described the basic components of an application architecture designed to support these types of resources in organisational settings. Down-stream processing and distribution options, support utilities, and the file structure used to organise projects, transcripts and related resources were also described. Interoperability between programs in the architecture and integration with third party tools was provided through the adoption of XML in an effort to 'future-proof' the design. By adopting an Open Source Approach, we benefited from the work of others and accelerated the development process. It also provided us with the opportunity to share our work with relevant research communities.

The near future involves ongoing refinement of the architecture's components. Currently the Digital Media Loop has been developed to provide processing features relevant for audio files only. Its functionality will be extended to support audio-visual transcription. Non-standard video feeds, including camera multiplexers and hemispherical cameras, will be of interest in our research projects. As a consequence Digital Video Processing software will likely be added to the architecture in Figure 3. The Indexer application utilises controls that display spoken language tagging taxonomies, which use system network notations. Conceptually these notations are suited to our applications but as they were not designed to be used as interactive controls further work is required in order to evaluate their usability. Usability testing will also be conducted on TransCoder to determine the best way of organising the interface to support different types of studies, and to support transcribers with various levels of proficiency and degrees of familiarity with the transcription process. The display capabilities of TransCoder will be eventually extended with security and privacy features.

## Acknowledgements

The research presented here was partially funded through two United Kingdom Engineering and Physical Sciences Research Council (EPSRC) grants, Semiotic Enterprise Design for IT Applications (SEDITA) jointly conducted by Staffordshire and Reading Universities, Grant Reference: GR/S04833/01, and Reducing Rework Through Decision Management (TRACKER) jointly conducted by Lancaster and Staffordshire Universities, Grant Reference: GR/R12176/01.